\documentclass[10pt,conference]{IEEEtran}
\IEEEoverridecommandlockouts
\usepackage{cite}
\usepackage{amsmath,amssymb,amsfonts}
\usepackage{algorithm}
\usepackage{algpseudocode}
\usepackage{amsmath}
\usepackage{graphicx}
\usepackage{caption}
\usepackage{fontawesome}
\usepackage{textcomp}
\usepackage{xcolor}  
\usepackage{float}
\usepackage{subcaption}

\usepackage[bookmarks=false]{hyperref}
\usepackage{colortbl} 
\usepackage{array} 
\usepackage{booktabs} 

\def\BibTeX{{\rm B\kern-.07em{\sc i\kern-.025em b}\kern-.08em
    T\kern-.1667em\lower.7ex\hbox{E}\kern-.125emX}}
\definecolor{tableHeader}{rgb}{0.6, 0.6, 0.9}
\definecolor{tableRow}{rgb}{0.9, 0.95, 1}
\definecolor{tableAltRow}{rgb}{0.8, 0.9, 1}
\definecolor{tableBorder}{rgb}{0.4, 0.6, 0.8}
\usepackage{graphicx}
\begin{document}

\title{FedMTFI: Feature Importance Based Optimized Multi Teacher Knowledge Distillation in Heterogeneous Federated Learning Environment}

\author{\IEEEauthorblockN{Nazmus Shakib Shadin, Aaron Cummings, Xinyue Zhang, and Bobin Deng}
\IEEEauthorblockA{\textit{Department of Computer Science, Kennesaw State University, Marietta, GA, 30060 USA}}}

\maketitle

\begin{abstract}
Federated learning (FL) is a decentralized approach that enables collaborative model training without exposing raw data. Instead of transferring sensitive data, it allows devices to share only model weights, keeping personal data locally and secure. However, in real world settings, the data held by devices is often not evenly distributed and devices mostly differ in computing power and memory capacity. These differences make FL harder to maintain consistent performance across the system. To address these issues, we propose FedMTFI, a novel architecture that combines multi-teacher knowledge distillation (MTKD) with feature importance to improve the FL process in heterogeneous environments. In FedMTFI, clients are clustered based on similar hardware and model types. Each cluster trains a specific model on not independently and identically distributed (non-IID) data. Within a cluster, every client updates that model using only its own local private data. The server then aggregates the locally trained models in each cluster using FedAvg to form multiple prototype models. Then these prototypes serve as teacher models to train a global generalized student model using MTKD. What makes FedMTFI more unique is the integration of Shapley values (SHAP) to emphasize important features during distillation, which enhances both accuracy and interpretability. Experimental results show that FedMTFI achieves higher accuracy than traditional FL algorithms and performs more effectively under non-IID data conditions.
\end{abstract}


\section{Introduction}
In today’s digital era, AI-powered edge devices are proliferating at an unprecedented rate, finding applications in smart homes \cite{guo2019review}, healthcare \cite{jiang2017artificial}, image processing \cite{he2016deep}, natural language processing \cite{tenney2019bert}, autonomous vehicles \cite{yurtsever2020survey}, etc. As these devices become more widespread and sophisticated, the need for real-time decision making, low latency, and robust privacy protection is more important than ever \cite{hamer2020fedboost, 9423193}. Federated learning (FL) has proven to be a promising solution to address these needs \cite{yang2019federated}. 
Unlike traditional centralized machine learning, which requires all data to be sent to a central server, FL keeps sensitive data on local devices and trains models directly where the data reside~\cite{caldas2018leaf}. Only the learned model weights or updates are sent to the server, preserving data privacy while enabling collaborative model training across devices~\cite{zhang2022energy}. One strategy for aggregating client weights in FL is FedAvg, which aggregates client-generated local model updates from clients by averaging them proportionally based on local dataset sizes \cite{beutel2020flower}. However, FedAvg assumes that all participating devices use the same model architecture and have comparable resources, which is an assumption that often fails in real world deployments \cite{li2021hermes}. In practice, client devices differ widely in computational power, memory, battery life, and connectivity \cite{wang2019interpret}. For example, a Raspberry Pi, a smartphone, and a laptop are unlikely to efficiently support the same deep learning model \cite{chen2023eefl, 9775925}. Applying a one-size-fits-all approach can overburden weaker devices and under-utilize more capable ones, leading to inefficiencies and poor performance~\cite{hamer2020fedboost}. This heterogeneity in device capabilities, model architectures, and data distribution is often referred to as model, device, and data heterogeneity respectively, which poses a significant challenge for FL systems~\cite{lin2020ensemble}. 

Therefore, there are efforts to develop novel FL frameworks suitable for heterogeneous client devices without sacrificing performance \cite{lin2020ensemble,li2020federated,li2019fedmd,hinton2015distilling}. Li et al. \cite{li2019fedmd} introduced a novel federated learning framework, FedMD, designed to support heterogeneous client models by leveraging knowledge distillation over a shared public dataset. The framework demonstrates an average accuracy improvement of approximately 20\% over standalone training and achieves performance close to centralized training benchmarks. Hinton et al. \cite{hinton2015distilling} proposed the knowledge distillation technique, where a smaller model learns from a larger model or ensemble using soft targets from a high-temperature softmax. This approach preserves generalization while reducing the model size. The results show that the student model retains more than 80\% of the ensemble’s performance gains on datasets like MNIST and speech recognition. In~\cite{li2020federated}, Li et al. extends FedAvg to better handle statistical and system heterogeneity in FL by introducing a proximal term to stabilize local updates or FedProx. It improves convergence and achieves significantly higher accuracy in highly heterogeneous environments. However, these works assume that all clients share the same local model and train the local model in the same way. 

Lin et al. \cite{lin2020ensemble} proposed an ensemble distillation framework for FL that supports heterogeneous models and non-IID data, using unlabeled or synthetic data for server-side distillation fuses client knowledge without requiring homogeneous models. The framework achieves higher accuracy and up to 3$\times$ fewer communication rounds than FedAvg and FedProx\cite{li2020federated}. Amirkhani et al. \cite{9522137} proposed a multi-teacher knowledge distillation (MTKD) framework for semantic segmentation, using five DABNet teachers trained on diverse datasets to guide a FastSCNN student via a score-weighted labeling system. By using diverse training scenarios, the method improves robustness to domain shift and noise, resulting in noticeably better performance than standard supervised learning on Cityscapes dataset. However, none of these literature incorporate interpretability into their design, which limits their effectiveness during global aggregation.

Our proposed method, FedMTFI, is a novel architecture designed for FL in heterogeneous environments where client devices differ in hardware capabilities, model structures, and local data distribution. FedMTFI is designed to embrace heterogeneity rather than fight it. Participating client devices are grouped based on their computational capability, which is then assigned a shared model architecture to train on local non-IID data. On the server side, model updates from all groups are aggregated into multiple prototype models using FedAvg. These prototypes act as teacher models, guiding a new global server model (student model) through MTKD~\cite{11401860}.

To further enhance the learning process, FedMTFI integrates Shapley values (SHAP) to identify and emphasize the most important features during MTKD. This ensures that the global student model not only learns from diverse teachers but also focuses on the most meaningful aspects of the data, improving both performance and interpretability\cite{shadin2025fedkdshap}. Through comprehensive testing in real world environments, our results show that FedMTFI consistently outperforms baseline approaches, and other well known FL techniques such as FedAvg, FedProx, which proves its effectiveness in handling the practical challenges of FL in resource-constrained and heterogeneous environments. In summary, the key contributions of this study are given below:
\begin{itemize}
    \item Firstly, we addressed real world FL challenges by grouping client devices based on model architecture and computational capabilities, enabling efficient training on non-IID data.
    \item Secondly, we introduced server-side MTKD, where multiple prototype models aggregated via FedAvg act as teacher models to guide the training of a server-side generalized model (student model).
    \item Finally, we integrated SHAP into the MTKD loss function to highlight important data features, improving model performance and interpretability.
\end{itemize}


\section{Literature Review}
\subsection{Device Heterogeneity in FL} Yang et al.~\cite{yang2021characterizing} present one of the first large-scale analyses of how device heterogeneity impacts FL. By examining data from 136,000 smartphones, the authors show that differences in device performance and availability slow training, reduce accuracy, and introduce unfairness. Notably, state heterogeneity (e.g., disconnections, user interruptions) has a greater impact than hardware differences alone. Even advanced FL methods like FedProx struggle under these conditions.
\vspace{-.05in}
\subsection{Knowledge Distillation for FL} Hinton et al.~\cite{hinton2015distilling} proposed knowledge distillation, where a smaller model learns from a larger model or ensemble model by using soft targets from the softmax. This approach preserves generalization while reducing the model size, with the student model retaining over 80\% of the ensemble's performance gains on datasets like MNIST. Building on this foundation, Li et al.~\cite{li2019fedmd} introduced FedMD, a federated learning framework that supports heterogeneous client models by leveraging knowledge distillation over a shared public dataset, which demonstrates approximately 20\% accuracy improvement over standalone training. Lin et al.~\cite{lin2020ensemble} further advanced this direction with an ensemble distillation framework that uses unlabeled or synthetic data for server-side distillation, achieving better and fewer communication rounds than FedAvg.
\vspace{-.05in}
\subsection{Multi-Teacher Knowledge Distillation} To address fairness under distribution shifts, Kenfack et al.~\cite{kenfack2024adaptive} proposed AGRE-KD, an adaptive ensemble distillation method that prioritizes the accuracy of the worst-case group by weighting teacher models based on gradient divergence from a biased reference. Experiments on benchmarks like Waterbirds and CelebA show that AGRE-KD achieves up to 91.9\% worst-case group accuracy. Amirkhani et al.~\cite{9522137} proposed a multi-teacher knowledge distillation framework for semantic segmentation, using five DABNet teachers trained on diverse datasets to guide a FastSCNN student via a score-weighted labeling system, which achieves up to 13.87\% mIoU gain on the Cityscapes dataset.
\vspace{-.05in}
\subsection{Shapley Values for FL Interpretability} Wang et al.~\cite{wang2019interpret} proposed a Shapley-value based method to interpret FL models while preserving privacy. The approach computes detailed feature importance for the host party's features and a unified Shapley score for guest features, preventing leakage of protected data. In a related work~\cite{wang2019measure}, the authors proposed a framework for quantifying participant contributions in both horizontal and vertical FL using deletion-based influence functions and Shapley values, enabling fair credit allocation without exposing private data.
\vspace{-.05in}
\section{Methodology}

\begin{figure*}[t]
    \centering
    \includegraphics[width=\textwidth]{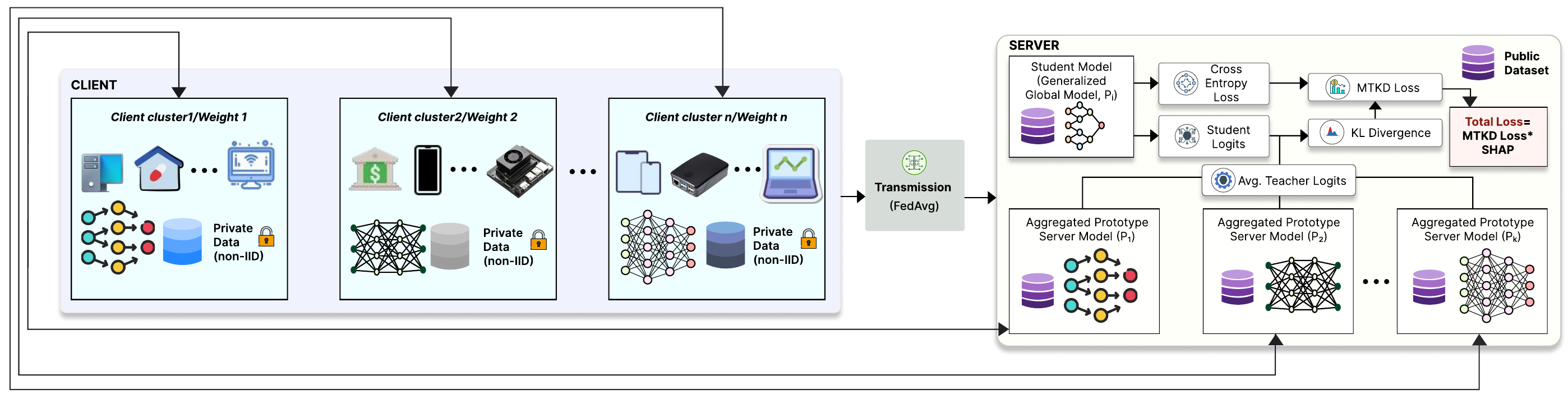}
    \captionsetup{font=footnotesize}
    \caption{Architectural Overview of FedMTFI Framework-The illustration of the proposed FedMTFI framework. Phase 1 (Client-Side Training): Heterogeneous client devices are grouped into hardware-specific clusters (e.g., CNN for resource-constrained devices, ResNet for capable devices), where each client trains locally on private, for simulating non-IID private data, we have partitioned the data via Dirichlet distribution. Phase 2 (Cluster Aggregation): Within each cluster, the server aggregates client model updates using FedAvg to produce cluster-specific prototype models that serve as teacher models for the post-hoc training stage. Phase 3 (Server-Side MTKD): Finally, the server performs post-hoc multi-teacher knowledge distillation on a public dataset, where each prototype acts as a teacher. The distillation combines cross-entropy loss on true labels and KL divergence between teacher ensemble and student predictions, weighted by SHAP-based feature importance scores ($\bar{\phi}$) to emphasize important features during knowledge transfer to the final global generalized model.}

    \label{figure:architecture}
\vspace{-.1in}
\end{figure*}

The proposed FedMTFI framework introduces a new FL workflow that combines cluster-based model aggregation, multi-teacher knowledge distillation, and SHAP-based feature weighting. In our implementation, we decompose the methodology into three components: client-end operations, server-end operations, and the transmission process. An overview of the system architecture is shown in Figure~\ref{figure:architecture}, which illustrates the interaction between clients and the server in each communication round. The following subsections describe each component of FedMTFI.

\begin{algorithm}[!hb]
\caption{FedMTFI: Federated Learning Process on the Client Side within a single cluster}
\label{alg:client}
\begin{algorithmic}[1]
\State \textbf{Input:} Client ID \( c \), global model weights for a specific cluster \( w_t \), local dataset \( (x, y)^{(c)} \), batch size \( b \), local epochs \( E \), balancing factor \( \alpha \), temperature \( T \), total no. of rounds \( N \)
\State \textbf{Output:} Final updated local weights \( w^{(c)} \)
\State\( \theta_S^{(c)} \leftarrow \texttt{Model\_n} \)
\State Partition private dataset using Dirichlet distribution with parameter \( \alpha \) across \( n \) clients.
\State Load client-specific partition: \( D^{(c)} = D^{\text{train}}_{(c)} \cup D^{\text{test}}_{(c)} \).

\For{each FL round \( r = 1 \) to \( N \)}
    \State \( \theta^{(c)}.\texttt{set\_weights}(w_t) \)
    \For{each batch \( (x_b, y_b) \in D^{\text{train}}_{(c)} \)}
        \State Compute logits: \( z = \theta^{(c)}(x_b) \)
        \State Compute loss: \( L = \text{CrossEntropy}(z, y_b) \)
        \State Perform backpropagation:
        \State \quad \( opt.\texttt{zero\_grad}() \)
        \State \quad \( L.\texttt{backward}() \)
        \State \quad \( opt.\texttt{step}() \)
    \EndFor
\EndFor
\State Evaluate on local test data:
\State \quad \( \text{loss}, \text{acc} \leftarrow \theta^{(c)}.\texttt{evaluate}(x_{\text{test}}, y_{\text{test}}) \)
\State Extract updated model weights:
\State \quad \( w^{(c)} \leftarrow \theta^{(c)}.\texttt{get\_weights}() \)
\end{algorithmic}
\end{algorithm}

\vspace{-.05in}
\subsection{Client-End Operation}
On the client side, each participating device (client) performs local training on its private data. 
The algorithm~\ref{alg:client} describes the training procedure inside a single federated learning round. 
At the beginning of round $r$, client $c$ receives the current global model parameters $w_t$ from the server. 
These parameters are used to initialize the client’s local model $\theta^{(c)}$. 
After the initialization, the client loads its private non-IID dataset $D^{(c)}$, which was partitioned using a Dirichlet distribution~\cite{li2019convergence} with parameter $\alpha$ to introduce statistical heterogeneity across clients. Lower values of $\alpha$ (e.g., $\alpha = 0.1$) result in highly heterogeneous data distributions, while higher values (e.g., $\alpha = 1.0$) produce more uniform partitions.

Once the setup is complete, the client trains $\theta^{(c)}$ for a fixed number of local epochs during round $r$. For each minibatch $(x_b, y_b)$ from the local training set $D^{(c)}_{\text{train}}$, the client performs a standard supervised learning step. The forward pass computes the model logits $z = \theta^{(c)}(x_b)$. 
A cross-entropy loss $L = \text{CrossEntropy}(z, y_b)$ is calculated using the ground truth labels. 
The optimizer then executes a gradient descent update on this loss. This process continues across every minibatch in the local epoch loop.

After completion of the local training phase for a specific round $r$, the client evaluates its updated model on the local test set $D^{(c)}_{\text{test}}$. This evaluation produces performance metrics such as accuracy and loss, which are used to monitor model behavior on the private data. Once the evaluation is completed, the client extracts its updated model parameters $w^{(c)} = \theta^{(c)}.\texttt{get\_weights()}$. These parameters, along with the evaluation metrics, are transmitted back to the central server.

The above procedure is repeated for each federated learning round $r = 1, 2, \dots, N$. By continuously training on private data and returning updated model parameters, each client contributes to improve the global model each prototype while retaining data privacy throughout the federated learning process.
\vspace{-.07in}
\subsection{Server-End Operation}
At the server side, FedMTFI introduces a cluster-based aggregation strategy combined with MTKD and SHAP-based contribution weighting. The algorithm~\ref{alg:server} provides a pseudocode summary of the server-side process. When the server receives model updates from the clients at the end of each round $r$, it first groups the client models by their respective cluster. Clients are partitioned into clusters based on characteristics such as device capability or model type. Each cluster has a homogeneous model architecture and similar capability. 
\begin{algorithm}[!hb]
\caption{FedMTFI: Federated Learning Process on the Server Side with Multi-Teacher Knowledge Distillation (MTKD) and Feature Importance (SHAP)}
\label{alg:server}
\begin{algorithmic}[1]
\State \textbf{Input:} Teacher models \( \theta_T^{(k)}\), Training dataset \( (x_{\text{train}}, y_{\text{train}}) \), Test dataset \( (x_{\text{test}}, y_{\text{test}}) \), Epochs \( E \), Balancing factor \( \alpha \), Temperature \( T \), Feature importance function \texttt{calculate\_feature\_importance()}, teacher\_logits \( \leftarrow [] \)
\State \textbf{Output:} Generalized global model (student model) \( \theta_S \)
\State \(x = x_{\text{train}}\) and \(y = y_{\text{train}}\)

\For{each \(\theta_T^{(k)}\), from \(1\) to \(K\) } 
    \State \( \theta_T^{(k)} \leftarrow \texttt{load\_model(k)} \)  
    \State \( \phi^{(k)} = \texttt{feature\_importance}(\theta_T^{(k)}, x, y) \)
    \EndFor
\State \( \bar{\phi} \leftarrow \frac{1}{K}(\sum_{k=1}^K \phi^{(k)})\)
\State \( \theta_S \leftarrow \texttt{create\_student\_model()} \)
\State \( \mathcal{D} \leftarrow \texttt{MultiTeacherDistiller}(\theta_S, \{ \theta_T^{(1)}, ..., \theta_T^{(K)} \}, \bar{\phi}) \)

\For{\( e = 1 \) to \( E \)}
  \For{\( (x, y) \in (x_{\text{train}}, y_{\text{train}}) \)}
    \State \( z_T^{(1)} = \theta_T^{(1)}(x) \), \( z_T^{(2)} = \theta_T^{(2)}(x) \)
    \For{ \( \theta_T^{(k)} \in \{\theta_T^{(1)}, ..., \theta_T^{(K)}\} \)}
        \State \( z_T^{(k)} \leftarrow \theta_T^{(k)}(x, \text{training=False}) \)
        \State Append \( z_T^{(k)} \) to teacher\_logits
    \EndFor
    \State \( Z_T \leftarrow \texttt{stack}(teacher\_logits, axis=0) \)
    \State \( \bar{z}_T \leftarrow \texttt{mean}(Z_T, axis=0)\)
    \State Calculate Student logits \( z_S = \theta_S(x) \)
    \State Compute soft targets \( \tilde{y}_T = \text{softmax}(\bar{z}_T / T) \), 
    \State \( \tilde{y}_S = \text{softmax}(z_S / T) \)
    \State \( L_{\text{total}} = (1 - \alpha) \cdot \text{CE}(z_S, y_b) + \alpha \cdot T^2 \cdot \text{KL}(\tilde{y}_T, \tilde{y}_S)
        \)
    \State \( \mathcal{L}_{\text{weighted}} = \bar{\phi} \cdot L_{\text{total}} \)
    \State Backpropagate and update \( \theta_S \)
  \EndFor
\EndFor

\State\( \text{loss}, \text{accuracy} \leftarrow \theta_S.\texttt{evaluate}(x_{\text{test}}, y_{\text{test}}) \)
\end{algorithmic}
\end{algorithm}
Within each cluster, the server performs the model aggregation. It applies FedAvg to the weights received from the clients of that cluster, which produces an aggregated cluster model. This cluster model is also referred to as a prototype model. Let $P_1, P_2, \dots, P_k$ denote each prototype model of the $k$ cluster. Each $P_k$ represents the consolidated knowledge of cluster $k$ after round $r$, and that will serve as an individual teacher in the next stage.

With the set of cluster models ready $\{P_1,\dots,P_k\}$, the server embarks on a MTKD process to synthesize a single generalized global model. The server initializes a new student model $\theta_s$ that will become the updated generalized global model. The server also loads the prototype model of each cluster to serve as a teacher model $\theta_T^{(k)} = P_k$. To effectively combine the knowledge of the teachers, the server uses a public or auxiliary dataset $(x_{\text{train}}, y_{\text{train}})$ that is representative of the overall task. This dataset can be unlabeled or labeled. In our implementation, we assume that a public dataset is available for the server to facilitate knowledge aggregation without accessing the private data of the clients.

Before training the server-side generalized model (final student model), the FedMTFI server computes the SHAP-based feature importance for each teacher model. Here, for each teacher model \( \theta_T^{(k)} \), a feature importance vector \( \phi^{(k)} \) is computed. Here we use the SHAP technique to quantify the contribution of each input feature to the teacher's predictions in the server dataset. Intuitively, $\phi^{(k)}_j$ measures how important the feature $j$ is to the model $\theta_T^{(k)}$ when producing the correct output. The server then derives a combined importance weight $\phi$ by aggregating the individual $\phi^{(k)}$. For example, taking an average between all $K$ teachers: $\phi = \frac{1}{K}\sum_{k=1}^K \phi^{(k)}$. The resulting $\phi$ captures the overall importance of the feature as agreed upon by the ensemble of teacher models. This information will be used to weight the knowledge distillation loss, ensuring that the distillation process pays more attention to features deemed important by the teachers.

The server now trains the generalized server model which is the MTKD based student model $\theta_s$ using ensemble MTKD. For each training, there is a fixed number of total epochs $E$, and the server iterates over mini-batches $(x_b, y_b)$ from the public training data. For a given batch, the server obtains predictions from all the teacher models and the student model as follows. It feeds the inputs $x_b$ into each teacher $\theta_T^{(k)}$ in inference mode to collect the teacher logits $z_T^{(k)} = \theta_T^{(k)}(x_b)$. The collection of logits from all teachers $\{z_T^{(1)}, z_T^{(2)}, \dots, z_T^{(K)}\}$ is then combined, by stacking and taking an average logits among teachers, to produce a ensemble teacher prediction $z_T = \frac{1}{K}\sum_{k=1}^K z_T^{(k)}$. In parallel, the student model produces its logits $z_S = \theta_s(x_b)$ for the same inputs. The server converts the averaged teacher logits into a probability distribution, which is also soft targets, $\tilde{y}_T = \mathrm{softmax}(z_T / T)$ using temperature $T$, and similarly $\tilde{y}_S = \mathrm{softmax}(z_S / T)$ for the student’s outputs. Then, it computes the MTKD loss between the student and the ensemble teacher and the standard loss on true labels and forms a total loss $L_{\text{total:}}$
\begin{equation}
   L_{\text{KD}} = D_{\mathrm{KL}}\!\big(\tilde{y}_T \,\|\, \tilde{y}_S\big), \qquad  
\end{equation}
\begin{equation}
    L_{\text{CE}} = \frac{1}{|y_b|}\sum_{i \in b} -\log P_{\theta_s}(y_i \mid x_i)\,,
\end{equation}
 
\begin{equation}
    L_{\text{total}} = (1-\alpha) L_{\text{CE}} + \alpha \, T^2 \, L_{\text{KD}}.
\end{equation}
Here, we use the factor $T^2$ with the Kullback-Leibler (KL) divergence term following standard practice in knowledge distillation to account for the temperature scaling. Finally, to incorporate feature importance, FedMTFI adjusts this loss by the Shapley-based weight. Since $L_{\text{total}}$ is a scalar loss value, we compute an aggregated feature importance scalar $\bar{\phi}$ by averaging the importance scores across all features:
\begin{equation}
    \bar{\phi} = \frac{1}{|F|} \sum_{i=1}^{|F|} \phi_i,
\end{equation}
where $|F|$ is the total number of features, and $\phi_i$ is the SHAP importance score for feature $i$. The weighted distillation loss is then computed as:
\begin{equation}
    L_{\text{weighted}} = \bar{\phi} \cdot L_{\text{total}}.
\end{equation}
This formulation ensures that batches with higher average feature importance contribute more to the learning process. The student model's parameters are then updated via backpropagation on $L_{\text{weighted}}$ for each batch.

The server repeats the above process for each batch and for $E$ epochs, which gradually refins the student model $\theta_s$ to align with the ensemble of teachers. After completion of the training, the generalized server global model (student model) $\theta_s$ learns from all cluster models (teacher models). It produces a distilled global model that is more robust and general. The server can evaluate this global model on a test set $(x_{\text{test}}, y_{\text{test}})$ to measure its accuracy and loss. By performing this cluster-based MTKD procedure, the server effectively leverages the knowledge distilled from multiple teacher models to build a unified global student model. This process ensures that insights from heterogeneous client groups are integrated into a single, high-performing model, addressing non-iid data disparities with model, and device heterogeneity in the FL process.

\textbf{Computational Overhead of SHAP:} While the exact Shapley value computation has exponential complexity $O(2^{|F|})$ for $|F|$ features, we employ a gradient-based SHAP approximation method~\cite{lundberg2017unified} specifically designed for deep neural networks. This approach leverages the model's gradients to efficiently approximate Shapley values by computing the expected gradients across a background distribution, reducing complexity to $O(B \cdot N)$, where $B$ is the background sample size and $N$ is the number of test samples. This gradient-based method is particularly suitable for our federated learning framework because it: (1) works effectively with complex deep learning models (e.g., CNNs) used in our cluster models, (2) provides accurate feature importance scores without requiring exponential computation, and (3) can be efficiently parallelized on the server's GPU. In our implementation, we have used a small background sample of 100 instances and computed SHAP values for one representative sample per class (10 samples total), making the overhead negligible compared to the overall training time. Furthermore, SHAP computation occurs only at the server side during the post-hoc distillation phase, and not during client-side local training, thereby preserving the communication efficiency of the federated learning process.

\vspace{-.07in}
\subsection{Transmission Process}

The FedMTFI framework follows a standard federated learning technique based on the FedAvg algorithm~\cite{li2021hermes}. Each client independently trains a local model and exchanges model parameters with the server within a cluster. At the beginning of each round, within a cluster, the server sends the current global model weights to the participating clients. After completion of a local training, each client sends its updated model weights and evaluation metrics back to the server. In our FedMTFI setup, communication between clients and the server is organized in a cluster-specific manner, enabling coordinated updates and aggregation within each cluster before integration at the global level. This design allows cluster-specific models to be updated and aggregated separately before being unified through the multi-teacher distillation process. This framework manages client selection and synchronization throughout communication rounds. After $N$ rounds, the system converges to a final global model that effectively captures and integrates knowledge from heterogeneous client clusters.

\section{Experimental Analysis}
To evaluate the effectiveness of FedMTFI, we have conducted experiments that capture key challenges in real-world federated learning environments. Specifically, we have analyzed convergence under different cluster and client settings, examined generalization on public datasets, and assessed the contribution of SHAP-based feature weighting through an ablation study. The client models are trained on non-IID MNIST, while the final distilled global model have been evaluated on FMNIST and CIFAR-10. We have also compared FedMTFI with FedAvg, FedProx, FedKDShap, and a centralized learning upper bound to provide a fair assessment of its performance. The FedMTFI source code is available on \faGithub~GitHub \footnote{\href{https://github.com/AIPO-Lab/FedMTFI}{https://github.com/AIPO-Lab/FedMTFI.}}.
\vspace{-.07in}
\subsection{Data Configurations}
\vspace{-.07in}
In our FedMTFI simulation, we used the MNIST, FMNIST, and CIFAR-10 datasets. The MNIST dataset, comprising 70,000 grayscale handwritten digit images of size 28×28 pixels across ten digit classes, was employed as private, non-IID training data distributed among client devices. To emulate realistic data heterogeneity, a Dirichlet partitioning strategy~\cite{li2019convergence} with parameter $\alpha$ was applied, creating statistically skewed data distributions across clients. This ensures that each client trains on a unique subset of MNIST digits, reflecting non-IID characteristics typical of real world federated environments.
For the evaluation of the global generalized model, we used FMNIST and CIFAR-10 as public auxiliary datasets. FMNIST includes 70,000 grayscale images (28×28 pixels) across ten fashion categories, such as sneakers, shirts, and dresses. CIFAR-10 consists of 60,000 color images (32×32 pixels) across ten everyday object categories, including airplanes, cats, and dogs. These public datasets enable server-side evaluation and assist validate the generalizability of the aggregated global model.
\vspace{-.07in}
\subsection{Simulation Settings}
\vspace{-.07in}
In the simulation of the proposed FedMTFI framework, each client trained locally using the Adam optimizer for 5 epochs over 30 communication rounds. We evaluated multiple combinations of cluster counts and participating clients, including settings with 3, 4, and 5 clusters and different numbers of active clients per round. This allowed us to observe how the variation of the participation of the clients and the configuration of the clusters impact the global loss and accuracy.
The cluster-based architecture employs four specialized model architectures to accommodate different device capabilities; for example: Cluster 0 (SimpleCNN): Lightweight CNN with 0.8M parameters optimized for resource-constrained devices.
Cluster 1 (ResNetLike): ResNet-inspired architecture with 1.5M parameters featuring residual blocks, balancing performance and efficiency.
Cluster 2 (MobileNetLike): MobileNet-inspired design with 1.2M parameters using depthwise separable convolutions for mobile and edge devices.
Cluster 3 (ResNet18Like): ResNet-18 inspired architecture with 2.1M parameters for computationally capable.

Additionally, a global StudentCNN model with 0.3M parameters serves as the MTKD target, aggregating knowledge from all cluster-specific teacher models through server-side distillation.

All architectures were designed with adaptive input channel support to handle both grayscale datasets (MNIST, FashionMNIST) and RGB datasets (CIFAR-10), ensuring compatibility across different data modalities while maintaining computational efficiency for heterogeneous federated environments.

\vspace{-.08in}
\subsection{Results}
We evaluated the proposed FedMTFI framework under multiple cluster and client configurations. Each client trained locally using the Adam optimizer for 5 epochs over 30 communication rounds. We report the global loss, global accuracy, and final global generalized student model performance on the FMNIST and CIFAR10 datasets. These metrics allow observation of convergence behavior, stability, and generalization across heterogeneous federated learning settings.
\subsubsection{Global Loss vs. Communication Rounds}
\begin{figure}[!ht]
    \centering
    \includegraphics[width=0.95\columnwidth]{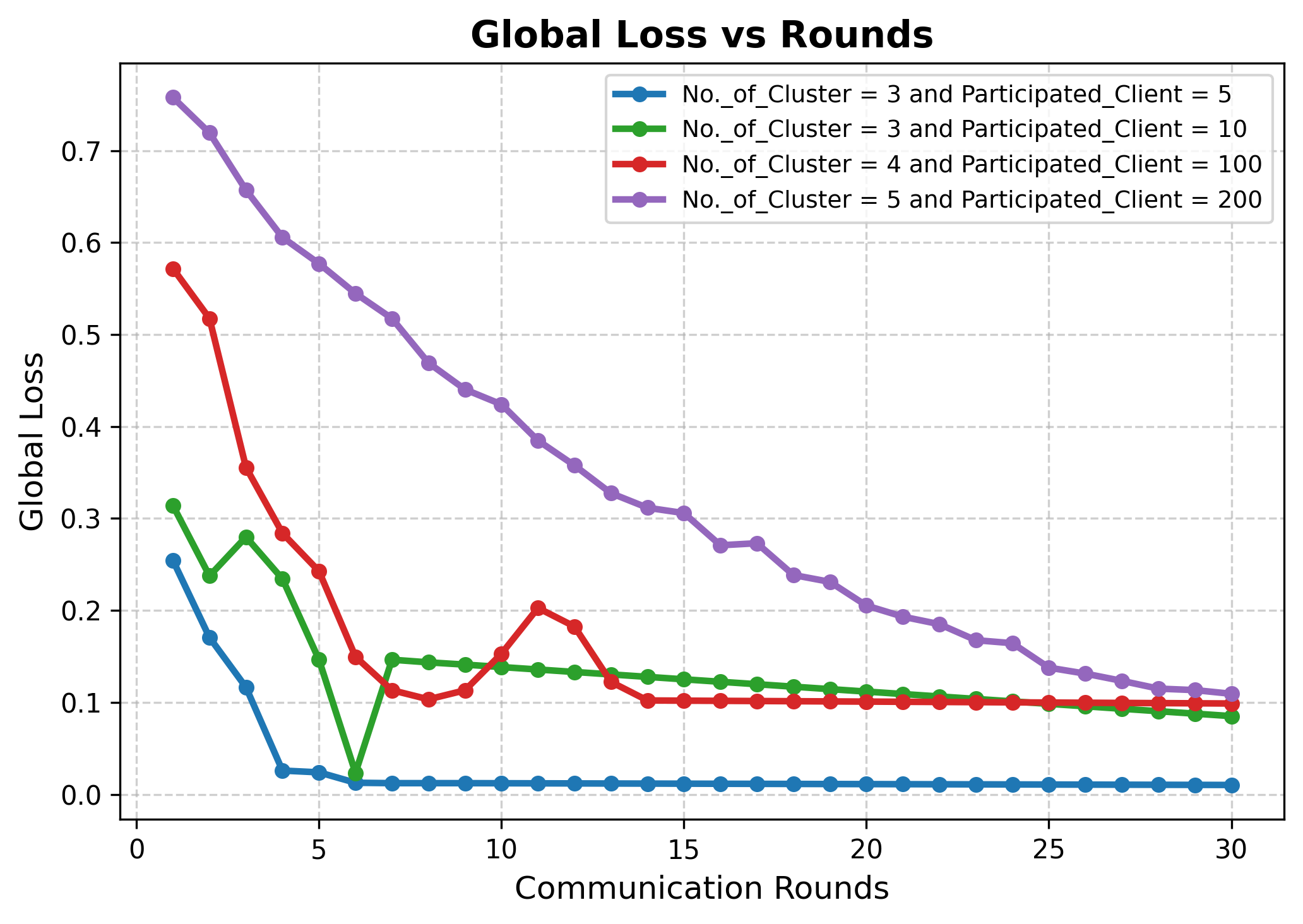}
    \captionsetup{font=footnotesize}
    \caption{FedMTFI: Global Loss comparison across clusters with varying client counts on MNIST private data.}
    \label{fig:client_models_loss}
\vspace{-.1in}
\end{figure}

The global loss decreased rapidly during the initial communication rounds for all configurations. As shown in Figure~\ref{fig:client_models_loss}, the configuration with three clusters and five clients achieved the fastest convergence, reaching a stable loss by round six. Configurations with higher client participation exhibited slower convergence due to increased variance among local updates, but remained stable after round ten. The configuration with four clusters and one hundred participating clients showed minor fluctuations before convergence.
\subsubsection{Global Accuracy vs. Communication Rounds}
\begin{figure}[!ht]
    \centering
    \includegraphics[width=0.95\columnwidth]{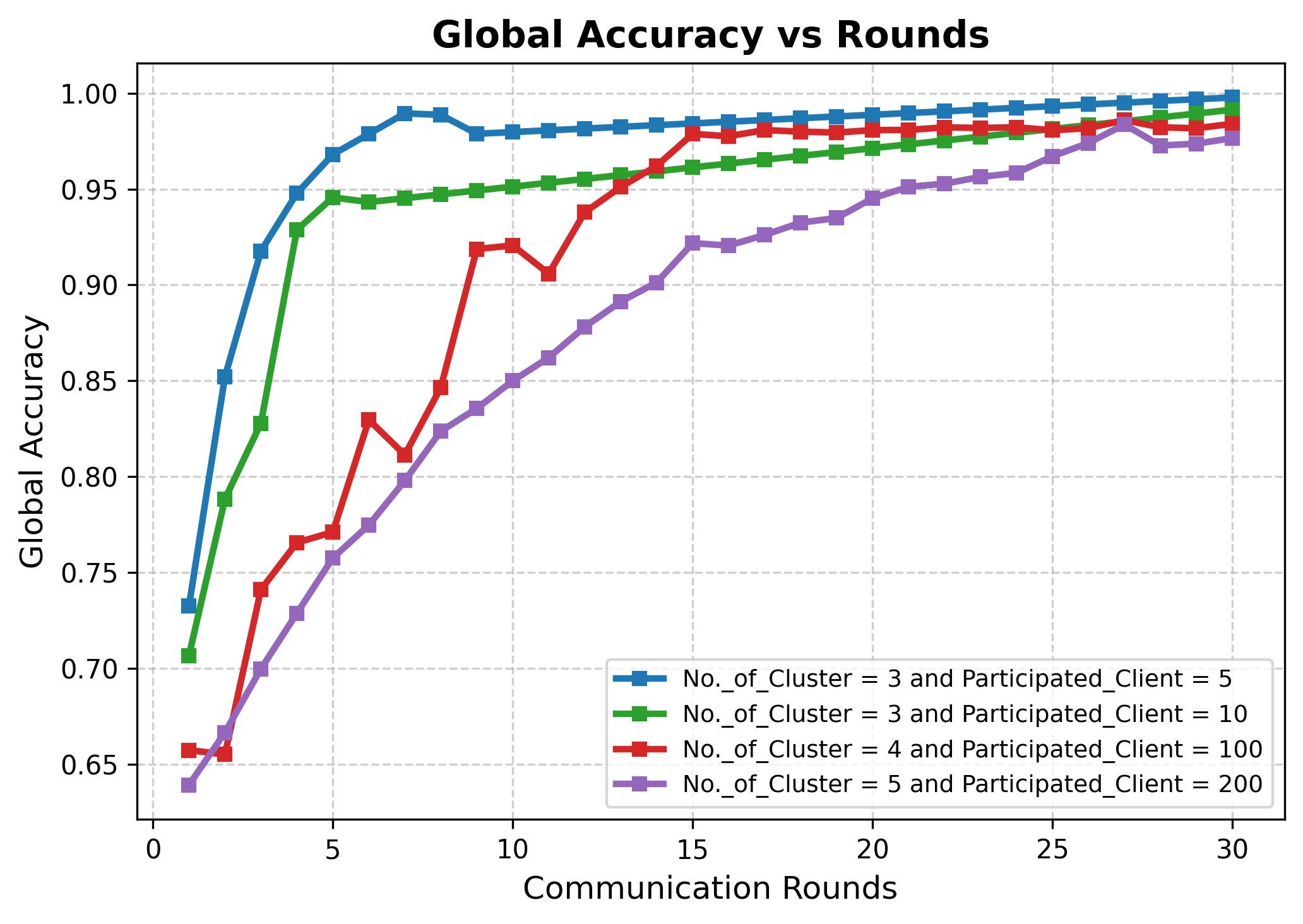}
    \captionsetup{font=footnotesize}
    \caption{FedMTFI: Global Accuracy comparison across clusters with varying client counts on MNIST private data.}
    \label{fig:global_accuracy}
\vspace{-.1in}
\end{figure}

Figure~\ref{fig:global_accuracy} shows that global accuracy improved consistently across all configurations. Three clusters with five clients achieved the fastest accuracy growth and produced the highest early-round accuracy. Three clusters with ten clients demonstrated moderate growth, while four clusters with one hundred clients required more rounds to converge due to increased data diversity. All configurations converged to comparable accuracy beyond round twenty.
\subsubsection{Sensitivity to Alpha ($\alpha$) Parameter}
To investigate the impact of data heterogeneity on client-side training performance, we conducted experiments with varying Dirichlet distribution parameter $\alpha$ applied to the private MNIST dataset distributed across clients. Lower $\alpha$ values create more heterogeneous (non-IID) data partitions, while higher values approach IID conditions. Table~\ref{table:alpha_sensitivity} presents the average client model accuracy on MNIST.

\begin{table}[!ht]
    \caption{Impact of Dirichlet $\alpha$ on Client Model Accuracy (MNIST)}
    \label{table:alpha_sensitivity}
    \centering
    \footnotesize
    \setlength{\tabcolsep}{12pt}
    \renewcommand{\arraystretch}{1.2}
    \begin{tabular}{|c|c|l|}
        \hline
        \rowcolor[HTML]{D3D3D3}
        \textbf{$\alpha$} & \textbf{MNIST Accuracy (\%)} & \textbf{Heterogeneity} \\
        \hline
        0.1 & 91.24 & Highly non-IID \\
        \rowcolor[HTML]{F2F2F2}
        0.3 & 93.18 & Moderately non-IID \\
        0.5 & 94.56 & Standard non-IID \\
        \rowcolor[HTML]{F2F2F2}
        1.0 & 96.12 & Mild non-IID \\
        10.0 & 97.85 & Near IID \\
        \hline
    \end{tabular}
\end{table}

The results show that client models maintain robust performance on MNIST across varying degrees of data heterogeneity. As expected, client-side accuracy improves with higher $\alpha$ values (less heterogeneous MNIST partitions), but even under extreme non-IID conditions ($\alpha = 0.1$), clients achieve competitive accuracy of 91.24\%. At $\alpha = 0.5$, the model achieves 94.56\% client accuracy, representing a realistic non-IID scenario commonly encountered in real world federated deployments.

\subsubsection{Final Generalized Global model Accuracy on FMNIST}
\begin{figure}[!ht]
    \centering
    \includegraphics[width=\columnwidth]{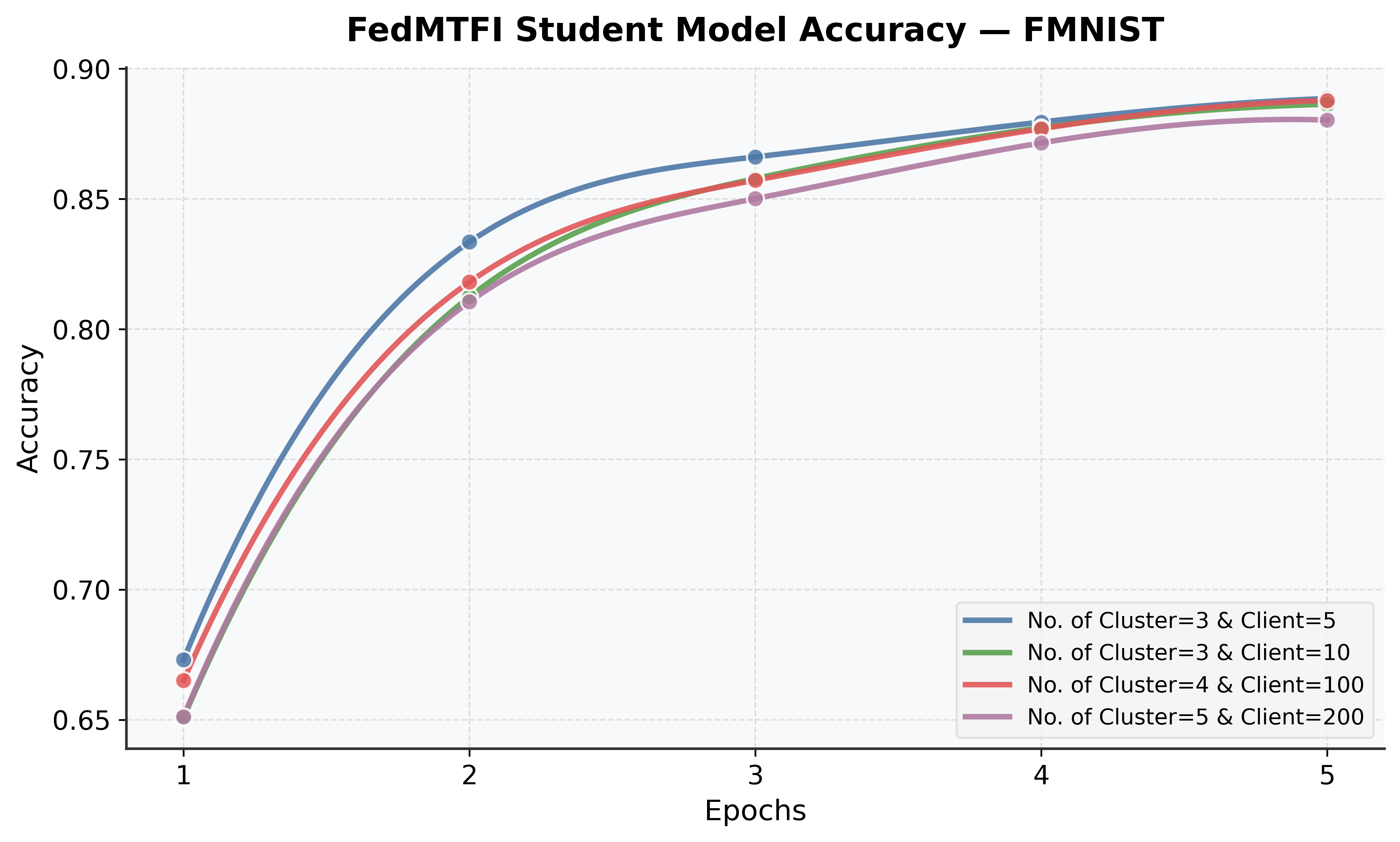}
    \captionsetup{font=footnotesize}
    \caption{FedMTFI: Final student model accuracy on FMNIST across epochs under varied cluster and client settings.}
    \label{fig:student_fmnist}
\vspace{-.1in}
\end{figure}

The performance of the final distilled generalized global model (student model) on FMNIST is shown in Figure~\ref{fig:student_fmnist}. All configurations exhibited consistent accuracy growth across epochs. The configuration with three clusters and five clients achieved the highest accuracy at the final epoch. Three clusters with ten clients and four clusters with one hundred clients demonstrated similar trends with slight variation in early epochs.
\subsubsection{Final Generalized Global model Accuracy on CIFAR10}
\begin{figure}[!hb]
    \centering
    \includegraphics[width=\columnwidth]{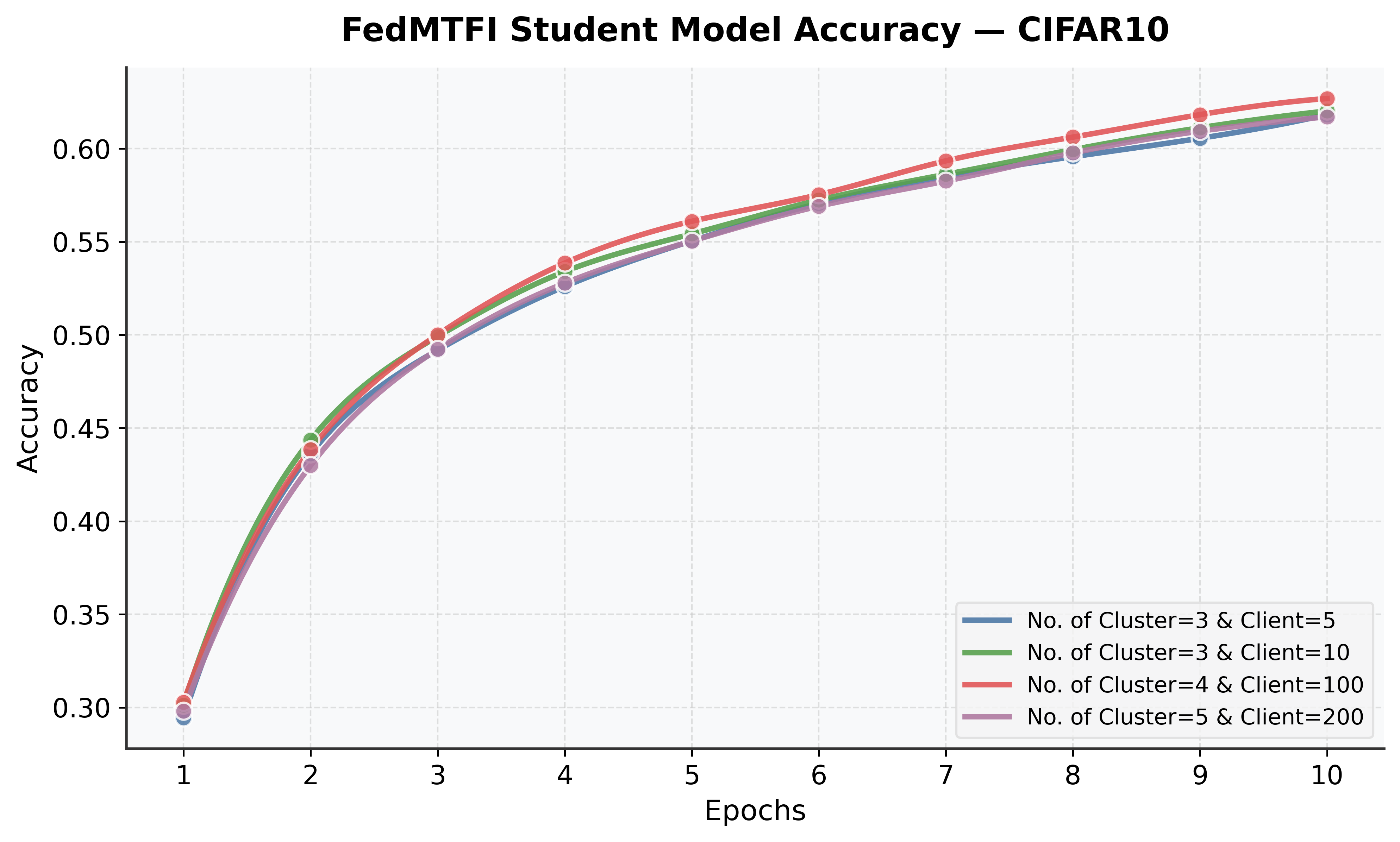}
    \captionsetup{font=footnotesize}
    \caption{FedMTFI: Final student model accuracy on CIFAR10 across epochs under varied cluster and client settings.}
    \label{fig:student_cifar}
\vspace{-.1in}
\end{figure}

Figure~\ref{fig:student_cifar} illustrates the final generalized global model performance on CIFAR10. Accuracy improved steadily across epochs for all configurations. The configuration with four clusters and one hundred clients achieved slightly higher accuracy in later epochs, which indicates improved generalization on more complex image distributions.
\subsection{Comparative Accuracy Analysis}
To evaluate the effectiveness of the proposed FedMTFI framework, a comparative analysis is conducted against few baseline federated learning algorithms and a centralized learning upper bound. The evaluation focuses on two benchmark datasets, CIFAR-10 and FMNIST, both commonly used for measuring generalization performance in image classification tasks.
\begin{table}[!ht]
    \caption{Accuracy Comparison of Centralized and Federated Learning Methods}
    \label{table:accuracy_fmnist_cifar10}
    \centering
    \footnotesize
    \setlength{\tabcolsep}{10pt}
    \renewcommand{\arraystretch}{1.2}
    \begin{tabular}{|l|c|c|c|c}
        \hline
        \rowcolor[HTML]{D3D3D3}
        \textbf{Method} & \textbf{CIFAR-10 (\%)} & \textbf{FMNIST (\%)} \\
        \hline
        Centralized Learning & 72.53 & 88.91 \\
        \rowcolor[HTML]{F2F2F2}
        FedAvg & 56.15 & 80.65 \\
        FedProx & 58.39 & 82.31 \\
        FedKDShap & 59.53 & 84.43 \\
        \rowcolor[HTML]{F2F2F2}
        FedMTFI (Proposed) & \textbf{64.48} & \textbf{87.28} \\
        \hline
    \end{tabular}
\end{table}
Table~\ref{table:accuracy_fmnist_cifar10} summarizes the results. Centralized learning serves as the upper bound (72.53\% on CIFAR-10, 88.91\% on FMNIST). Among federated baselines, FedAvg achieved 56.15\% and 80.65\%, while FedProx~\cite{li2020federated} improved to 58.39\% and 82.31\% through regularization for client heterogeneity. FedKDShap~\cite{shadin2025fedkdshap}, which integrates Shapley values with knowledge distillation to identify important features for guiding student model training on non-IID data, achieved 59.53\% and 84.43\%. The proposed FedMTFI outperformed all baselines, which reaches 64.48\% on CIFAR-10 and 87.28\% on FMNIST. This improvement demonstrates the effectiveness of combining MTKD with cluster-specific model refinement. The results show that FedMTFI narrows the gap between federated and centralized learning by improving feature alignment and reducing the adverse effects of heterogeneous data distributions. Overall, FedMTFI's results show that adding interpretability-guided MTKD and cluster-based aggregation improves accuracy under non-IID data on multiple datasets.

\subsection{Ablation Study: Effect of SHAP Weighting}
To separate the contribution of SHAP-based feature importance weighting, we have conducted an ablation study comparing FedMTFI with SHAP weighting ($L_{\text{weighted}}$) against FedMTFI without SHAP weighting (using $L_{\text{total}}$ directly). Table~\ref{table:ablation_shap} presents the results in the configuration of 3 clusters with 10 clients over 30 rounds.

\begin{table}[!ht]
    \caption{Ablation Study: Effect of SHAP Weighting on FedMTFI}
    \label{table:ablation_shap}
    \centering
    \footnotesize
    \setlength{\tabcolsep}{10pt}
    \renewcommand{\arraystretch}{1.2}
    \begin{tabular}{|l|c|c|}
        \hline
        \rowcolor[HTML]{D3D3D3}
        \textbf{Configuration} & \textbf{CIFAR-10 (\%)} & \textbf{FMNIST (\%)} \\
        \hline
        FedMTFI w/o SHAP ($L_{\text{total}}$) & 67.94 & 91.55 \\
        \rowcolor[HTML]{F2F2F2}
        FedMTFI w/ SHAP ($L_{\text{weighted}}$) & \textbf{70.61} & \textbf{93.43} \\
        \hline
        \textbf{Improvement} & +2.67 & +1.88 \\
        \hline
    \end{tabular}
\end{table}

The results demonstrate that the incorporation of SHAP-based feature importance weighting provides consistent accuracy improvements of +2.67\% on CIFAR-10 and +1.88\% on FMNIST. This improvement validates the effectiveness of the feature importance weighting mechanism in guiding the distillation process to focus on more informative features. The ablation shows that SHAP weighting is a meaningful contributor to FedMTFI's performance gains beyond the MTKD alone. In future work, we will investigate comparisons with more recent FL methods such as FedDF~\cite{lin2020ensemble} and SCAFFOLD~\cite{karimireddy2020scaffold} to further validate the generalizability of FedMTFI across various FL scenarios.

\vspace{-.07in}
\section{Conclusion}
This paper presented FedMTFI, a novel federated learning framework designed to tackle the challenges of device, model, and data heterogeneity in edge computing environments where privacy and the utilization of training capacity are important. By integrating multi-teacher knowledge distillation (MTKD) with Shapley value (SHAP)-based feature importance, FedMTFI enables the aggregation of knowledge from diverse client architectures into a single, robust global model. Our extensive experiments on MNIST, FMNIST, and CIFAR-10 datasets demonstrate that FedMTFI consistently outperforms traditional federated learning algorithms, such as FedAvg and FedProx, particularly under non-IID data conditions. The proposed cluster-based aggregation strategy allows clients with varying computational capabilities to participate effectively, while the SHAP-guided distillation ensures that the global model prioritizes the most informative features, thereby enhancing both learning efficiency and model interpretability. Despite these advantages, the integration of SHAP-based feature importance introduces a computational trade-off. The calculation of Shapley values requires additional forward passes during the server-side distillation phase, which can lead to an increased processing time compared to standard aggregation methods. However, we argue that this overhead is justified by the significant gains in model accuracy and convergence speed, especially in scenarios where communication bandwidth is a scarcer resource than server-side computation and client devices vary in training capacity. Furthermore, the use of gradient-based SHAP approximations effectively mitigates the exponential complexity of exact Shapley value computation, making the framework feasible for practical deployment.

Finally, FedMTFI offers a promising pathway for scalable and privacy-preserving decentralized learning across heterogeneous real world edge devices. The framework not only democratizes participation in federated learning by accommodating diverse hardware, but also produces global models that are both accurate and interpretable. Future research will focus on two key directions: first, optimizing the SHAP estimation process to further reduce computational costs without compromising feature importance accuracy; and second, extending the FedMTFI framework to more complex and dynamically changing federated environments, including those with time varying data distributions and intermittent client availability.
\section*{Acknowledgments}
The work is partially supported by the U.S. National Science Foundation under grants NSF-2348417 and NSF-2431597.

\bibliographystyle{IEEEtran}
\bibliography{ref.bib}
\end{document}